\documentclass[runningheads]{llncs}
\usepackage{graphicx}
\usepackage[unicode]{hyperref}
\usepackage[ruled,vlined,linesnumbered]{algorithm2e}
\SetKwRepeat{Do}{do}{while}

\begin{document}
\title{DPLL(MAPF): an Integration of Multi-Agent Path Finding and SAT Solving Technologies}

%
\titlerunning{DPLL(MAPF)}
%
\author{Martin Čapek\inst{1} \and Pavel Surynek\inst{1}\orcidID{0000-0001-7200-0542}}
%
\authorrunning{Martin Čapek and Pavel Surynek}
%
\institute{Faculty of Information Technology, Czech Technical University in Prague\\Th\'{a}kurova 9, 160 00 Praha 6, Czechia\\
\email{\{capekma9,pavel.surynek\}@fit.cvut.cz}}
\maketitle              
\begin{abstract}
In multi-agent path finding (MAPF), the task is to find non-conflicting paths for multiple agents from their initial positions to given individual goal positions. MAPF represents a classical artificial intelligence problem often addressed by heuristic-search. An important alternative to search-based techniques is compilation of MAPF to a different formalism such as Boolean satisfiability (SAT). Contemporary SAT-based approaches to MAPF regard the SAT solver as an external tool whose task is to return an assignment of all decision variables of a Boolean model of input MAPF. We present in this short paper a novel compilation scheme called DPLL(MAPF) in which the consistency checking of partial assignments of decision variables with respect to the MAPF rules is integrated directly into the SAT solver. This scheme allows for far more automated compilation where the SAT solver and the consistency checking procedure work together simultaneously to create the Boolean model and to search for its satisfying assignment.

\end{abstract}
\section{Introduction}
Multi-agent path finding (MAPF) \cite{DBLP:conf/focs/KornhauserMS84,DBLP:journals/jair/Ryan08,DBLP:conf/aiide/Silver05,DBLP:journals/ai/SharonSGF13,DBLP:journals/ai/SharonSFS15,DBLP:conf/icra/Surynek09,DBLP:journals/jair/WangB11} is the task of navigating agents from given starting positions to given individual goal positions. The task takes place in an undirected graph $G=(V,E)$. Agents from a set $A=\{a_1,a_2,...,a_k\}$ are located in vertices of $G$ with at most one agent per vertex. Movements of agents are instantaneous and are possible across edges assuming no other agent is entering the same target vertex. Agents are allowed to enter vertices being simultaneously vacated by other agents. However, the trivial case when a pair of agents swaps their positions across an edge is forbidden in the standard formulation of MAPF.

We usually denote the configuration of agents in vertices of $G$ at a discrete time step $t$ as $\alpha_t: A \rightarrow V$. Non-conflicting movements transform configuration $\alpha_t$ {\em instantaneously} into the next configuration  $\alpha_{t+1}$. We do not consider what happens between $t$ and $t+1$ in this discrete abstraction. Multiple agents can move at a time hence the MAPF problem is inherently parallel.

The initial configuration of agents in vertices of the graph can be written as $\alpha_0: A \rightarrow V$ and similarly the goal configuration as $\alpha_+: A \rightarrow V$. The task of navigating agents hence can be expressed as a task of transforming the initial configuration of agents $\alpha_0: A \rightarrow V$ into the goal configuration $\alpha_+: A \rightarrow V$ via valid moves.

The MAPF problem represents an important abstraction for many real-life tasks from robotics \cite{DBLP:journals/corr/abs-1810-03071}, warehouse logistics \cite{DBLP:journals/tase/BasileCC12}, computer games \cite{DBLP:conf/iccbr/WenderW14}. In order to reflect various aspects of real-life applications, variants of MAPF have been introduced such as those considering {\em kinematic constraints} \cite{DBLP:conf/ijcai/HonigK00XAK17}, {\em large agents} \cite{DBLP:conf/aaai/LiSF0KK19}, or {\em deadlines} \cite{DBLP:conf/ijcai/0001WFLKK18} - see \cite{DBLP:journals/corr/0001KA0HKUXTS17} for more variants.

\subsection{Compilation of MAPF to SAT}

Compiling MAPF to other formalism for which an off-the-shelf solver is available is a popular solving approach. Optimal solvers for MAPF based on the compilation to {\em constraint satisfaction problem} (CSP) \cite{DBLP:conf/ausai/Ryan08}, {\em answer set programming} (ASP) \cite{DBLP:journals/tplp/Bogatarkan020}, integer programming (IP) \cite{DBLP:conf/ijcai/LamBHS19}, and Boolean satisfiability (SAT) \cite{DBLP:journals/amai/Surynek17} currently exist.

In this paper, we focus on compilation of MAPF to SAT \cite{Biere:2009:HSV:1550723}. Contemporary techniques of MAPF compilation regard the SAT solver as an external tool having only limited interaction with the main MAPF solver. Often, a formula, called a {\em complete Boolean model}, encoding a question whether there exists a solution to input MAPF of a specified cost is constructed in a single-shot and consulted with the SAT solver \cite{DBLP:journals/amai/Surynek17}. The task of the SAT solver is to determine the truth value assignment of all decision variables satisfying the formula or answer that such an assignment does not exist. This scheme has been used in MDD-SAT, the first sum-of-costs optimal SAT-based MAPF solver \cite{DBLP:conf/ecai/SurynekFSB16}.

The disadvantage of the SAT consultation scheme from MDD-SAT is twofold: {\bf (1)} the complete Boolean model must be fully specified so that the equivalence between the solvability of the input MAPF instance and satisfiability of the Boolean model is established which may result in a large formula, and {\bf (2)} the SAT solver in this scheme acts as a black box for the main solver that has no way interact with the SAT solver until it finishes.

The first disadvantage has been addressed in SMT-CBS \cite{DBLP:conf/ijcai/Surynek19}, a sum-of-costs optimal SAT-based solver that introduced a concept of {\em incomplete Boolean models}. Using the incomplete Boolean model, the input MAPF instance is not fully specified so only the implication between the solvability of the input MAPF and the Boolean model holds. Such a relaxation requires that the MAPF solution obtained the from truth value assignment of the model is checked for consistency against MAPF movement rules (as those are not fully encoded in the model). If the rules are not violated, the solving process is finished. Otherwise the model needs to be refined by constraints that forbid the detected MAPF rule violations and consulted with the SAT solver again \footnote{This process is analogous to conflict-based search (CBS) \cite{DBLP:journals/ai/SharonSFS15} where MAPF rule violations (conflicts between pairs of agents) are resolved via branching the search.}.

The benefit of using incomplete Boolean models is that often the solving process finishes with a small formula since lot of constraints specifying the MAPF problem completely will not come into effect (for example we do not need to specify all collision avoidance constraints between agents in a sparse instance). A similar process has been adopted in compilation of MAPF to IP \cite{DBLP:conf/ijcai/LamBHS19}.

However, the second disadvantage only becomes more apparent in SMT-CBS. The main MAPF solver must wait until the complete truth value assignment is found by the SAT solver even if the SAT solver makes an early decision leading to MAPF rule violation which however the SAT solver cannot detect because it is not aware of the rules and the communication with the main solver that knows the rules is absent.

\subsection{Contribution}
We contribute by introducing a concept of turning a SAT solver into a MAPF solver via closer integration with the main MAPF solving part. Instead of using the SAT solver as a black box we suggest to extend it with understanding of the MAPF rules so it can check if the current partial assignment violates the MAPF rules or not at any time during the search for complete satisfying truth value assignment. If a violation of MAPF rules from the partial assignment is detected, the consistency checking procedure immediately suggests a model refinement via adding new constraints (clauses) \footnote{New decision variables can be added as well, but adding new variables is not needed for the standard variant of MAPF.} and the search continues with no interruption. Such a scheme allows for simultaneous construction of an incomplete Boolean model of MAPF together with searching for its satisfying assignment.

The SAT solving technology and related {\em satisfiability modulo theories} (SMT) offer concepts that can be utilized in the suggested scheme, namely DPLL($T$) \cite{DBLP:journals/jacm/NieuwenhuisOT06,DBLP:conf/fmcad/KatzBTRH16} which is a framework for integrating the SAT solver with a decision procedure, usually denoted $\mathit{DECIDE}_T$, for the conjunctive fragment of some first-order theory $T$. The two components of DPLL($T$) together form a decision procedure for general problems in theory $T$ with arbitrary Boolean structure where the SAT solver component takes care of the Boolean structure and the $\mathit{DECIDE}_T$ component checks the consistency of assignments suggested by the SAT solver against axioms of $T$.

In the rest of the paper, we show how to build DPLL(MAPF), that is DPLL($T$) where theory $T$ is substituted by a theory defining MAPF movement rules. DPLL(MAPF) will be introduced in 3 steps, as a gradual refinements of MDD-SAT and SMT-CBS.
    
\section{Background}

\begin{figure}[h]
  \centering
  \includegraphics[width=0.75\linewidth]{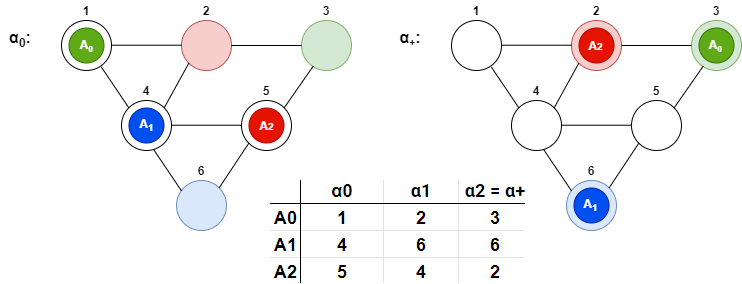}
  \caption{An example of MAPF consisting of three agents and its solution.}
  \label{fig-MAPF}
\end{figure}

An instance of MAPF will be denoted $\Sigma = (G=(V,E),A,\alpha_0,\alpha_+))$ (see Figure \ref{fig-MAPF}). We often aim on minimizing global cumulative cost functions like {\em makespan}, the total time until the last agent reaches its goal position, or {\em sum-of-costs} the total number of move/wait actions of agents before reaching their goal positions. While finding a feasible solution of MAPF can be done in polynomial time \cite{DBLP:conf/ijcai/LunaB11,DBLP:conf/atal/WildeMW13}, finding an optimal solution with respect to either the makespan or the sum-of-costs is an NP-hard problem \cite{DBLP:conf/aaai/RatnerW86}.

\subsection{Eager Encoding: MDD-SAT}

The idea behind MDD-SAT \cite{DBLP:conf/ecai/SurynekFSB16} is to construct a {\em complete Boolean model}, a propositional formula $\mathcal{F}(\xi)$ according to the following definition.

\begin{definition}
  {\bf (complete model).} Propositional formula $\mathcal{F(\xi)}$ is a {\em complete Boolean model} for MAPF $\Sigma$
  iff $\mathcal{F(\xi)}$ is satisfiable $\Leftrightarrow$ $\Sigma$ has a solution of sum-of-costs $\xi$.
\end{definition}

Being able to construct $\mathcal{F}(\xi)$ for solvable MAPF, one can obtain the optimal sum-of-costs by consulting the SAT solver with a series of queries about $\mathcal{F}(\xi_0)$, $\mathcal{F}(\xi_0 + 1)$, ... until a satisfiable formula is found, where $\xi_0$ is a lower bound on the sum-of-costs calculated as the sum of shortest paths of individual agents. This iterative scheme works due to the fact that satisfiability of $\mathcal{F}(\xi)$ is a non-decreasing function in parameter $\xi$.

\begin{figure}[h]
 \centering
\includegraphics[width=0.45\linewidth]{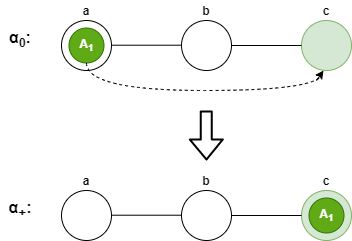}
  \caption{Example of MAPF with one agent}
  \label{fig-PATH}
\end{figure}

\begin{figure}[h]
 \centering
  \includegraphics[width=0.7\linewidth]{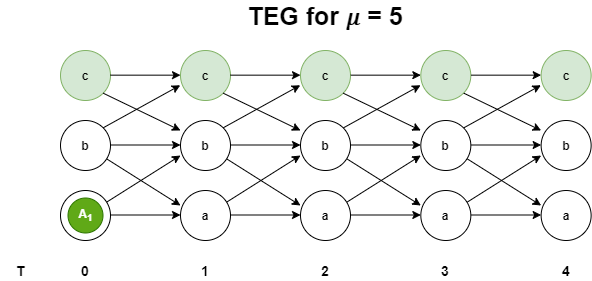}
 \caption{A TEG for MAPF from figure \ref{fig-PATH}.}
 \label{fig-TEG}
\end{figure}

The construction of $\mathcal{F}(\xi)$ must ensure that a valid MAPF solution can be extracted from its satisfying assignment. This is done by representing configurations of agents at all relevant time steps before they reach their goals via propositional variables. We first make a {\em time expanded graph} (TEG) of the underlying graph $G$ \cite{DBLP:journals/amai/Surynek17} for each agent, a directed acyclic graph obtained by copying vertices of $G$ for all relevant time steps. A directed edge is introduced into TEG for each pair of nodes from consecutive copies corresponding to vertices that are connected in $G$. In addition to this, nodes from consecutive copies corresponding to identical vertices are connected by directed edges as well to represent wait actions. A directed path in TEG corresponds to an individual plan of agent (sequence of its moves). The construction of TEG is shown in Figures \ref{fig-PATH} and \ref{fig-TEG}.

It remains to ensure that satisfying assignments of $\mathcal{F}(\xi)$ correspond to non-conflicting paths. A Boolean variable $\mathcal{X}_u^t(a_i)$ is introduced for every node $u_t$ from TEG (a node corresponding to $u \in V$ at time step $t$) for each agent agent $a_i \in A$; $\mathcal{X}_u^t(a_i)$ is $\mathit{TRUE}$ iff agent $a_i$ is located in $u$ at time step $t$. Similarly, we introduce Boolean variables for edges denoted $\mathcal{E}_{u,v}^t(a_i)$, with analogous meaning; $\mathcal{E}_{u,v}^t(a_i)$ is $\mathit{TRUE}$ iff agent $a_i$ moves from $u$ to $v$ starting the move at time step $t$. Finally constraints are added so that truth assignments are restricted to those that correspond to valid solutions of a given MAPF. The added constraints together ensure that $\mathcal{F(\xi)}$ is a {\em complete Boolean model} for given MAPF. We omit the detailed list of the constraints and refer the reader to \cite{DBLP:conf/ecai/SurynekFSB16}.

The MDD-SAT solver implements an improvement over TEGs based on the observation that not all nodes in TEG can be reached under given sum-of-costs $\xi$. The unreachable nodes can be pruned out from the TEG resulting in a directed acyclic graph called {\em multi-value decision diagram} (MDD) which has been introduced in the context of MAPF by search-based solvers \cite{DBLP:journals/ai/SharonSGF13,DBLP:journals/ai/SharonSFS15}. Adoption of MDDs in MDD-SAT resulted in much smaller formulae.

	    

\subsection{Lazy Encoding: SMT-CBS}

An important innovation step from MDD-SAT is represented by SMT-CBS \cite{DBLP:conf/ijcai/Surynek19}, an optimal SAT-based solver employing the idea of encoding MAPF as a Boolean formula {\bf lazily}. The lazy encoding is formalized through the concept of {\em incomplete Boolean model} defined as follows.

\begin{definition}
  {\bf (incomplete model).} Propositional formula $\mathcal{H(\xi)}$ is an {\em incomplete Boolean model} of MAPF 
  $\Sigma$ iff $\mathcal{H(\xi)}$ is satisfiable $\Leftarrow$ $\Sigma$ has a solution of sum-of-costs $\xi$.
\end{definition}

In an incomplete Boolean model $H(\xi)$ we do not specify all constraints defining the movement rules of MAPF. We rely on being lucky to obtain a valid MAPF solution from an under-specified formulation. Hence, in contract to MDD-SAT, we need to add a check that the solution extracted from the satisfying assignment of $\mathcal{H(\xi)}$ is consistent, that is, we need to ensure that agents do not jump, do not disappear, do not appear from nothing etc. since the correspondence between non-conflicting directed paths in MDDs and satisfying assignments of $H(\xi)$ is no longer preserved in the under-specified formulation. If the consistency check is successfully passed, we can return the valid MAPF solution extracted from the model otherwise the incomplete model needs to be refined.


\begin{algorithm}[h]
\begin{footnotesize}
\SetKwBlock{NRICL}{SMT-CBS ($\Sigma = (G=(V,E),A,\alpha_0,\alpha_+))$}{end} \NRICL{
    $\mathit{conflicts} \gets \emptyset$\\
    $\mathit{paths} \gets$ $\{$shortest path from $\alpha_0(a_i)$ to $\alpha_+(a_i) | i = 1,2,...,k\}$ \\   
    $\xi \gets \sum_{i=1}^k{\xi(\mathit{paths}(a_i))}$ \\
    \While {$\mathit{TRUE}$}{
         $(\mathit{paths,conflicts}) \gets$ SMT-CBS-Fixed($\mathit{conflicts},\xi,\Sigma$)\\
        \If {$\mathit{paths} \neq$ UNSAT}{
        	\Return $\mathit{paths}$\\
        }
        $\xi \gets \xi + 1$\\
    }
}   
 
\SetKwBlock{NRICL}{SMT-CBS-Fixed($conflicts,\xi,\Sigma$)}{end} \NRICL{
	    $\mathcal{H}(\xi) \gets$ encode-Incomplete$(\mathit{conflicts},\xi,\Sigma)$\\
	    \While {$\mathit{TRUE}$}{
	        $\mathit{assignment} \gets$ consult-SAT-Solver$(\mathcal{H}(\xi))$\\
	        \If {$\mathit{assignment} \neq \mathit{UNSAT}$}{
	            $\mathit{paths} \gets$ extract-Solution$(\mathit{assignment})$\\
	            $\mathit{collisions} \gets$ check-Consistency($\mathit{paths}$) /* via $\mathit{DECIDE}_{\mathit{MAPF}}$ */\\
                   \If {$\mathit{collisions} = \emptyset$}{
                      \Return $(\mathit{paths,conflicts})$\\
                   }
                   \For{each $(a_i,a_j,v,t) \in \mathit{collisions}$}{
                      $\mathcal{H}(\xi) \gets \mathcal{H}(\xi) \cup \{\neg \mathcal{X}_v^t(a_i) \vee \neg \mathcal{X}_v^t(a_j)$\}\\
                      $\mathit{conflicts} \gets \mathit{conflicts} \cup \{[(a_i,v,t),(a_j,v,t)]\}$
                   }
               }
               \Return {(UNSAT,$\mathit{conflicts}$)}\\
          }
}
\caption{SMT-CBS algorithm for MAPF solving} \label{alg-SMT-CBS}
\end{footnotesize}
\end{algorithm}

The pseudo-code of SMT-CBS algorithm is shown as Algorithm \ref{alg-SMT-CBS}. The high-level loop that iterates sum-of-costs is the same as in MDD-SAT. The difference rests in low-level loop within the SMT-CBS-Fixed function that answers existence of a solution of specified sum-of-costs $\xi$ in which the incomplete Boolean model is refined. Various strategies of refinements can be adopted. The SMT-CBS starts with $H(\xi)$ where only collision avoidance constraints are omitted. The constraints making agents to move along directed paths in MDDs are present. Hence the consistency check consists in a check for collisions (lines 16-18). If a collision in a vertex is detected, say a collision between agents $a_i$ and $a_j$ in $v \in V$ at time step $t$, then the model is refined with collision avoidance constraints, in this case clause $\neg \mathcal{X}_v^t(a_i) \vee \neg \mathcal{X}_v^t(a_j)$ to $H(\xi)$ is added (line 20). Eventually $H(\xi)$ may converge towards the complete Boolean model however often a solution is obtained much earlier.

\section{SAT Solver + MAPF = DPLL(MAPF)}

The SAT solver in the SMT-CBS framework is put in position that it does not completely understand what is the problem being solved. It may happen that early variable assignments in the SAT solver's search are inconsistent with MAPF rules. However the MAPF solution consistency check can be made only after the SAT solver assigns all variables. Hence we suggest to make a further innovation step from SMT-CBS and to check consistency also in partial assignments of incomplete Boolean models. This step requires very close integration of the SAT solver and the MAPF part.

The pseudo-code of the low-level search of DPLL(MAPF) that checks existence of MAPF solution for fixed sum-of-costs $\xi$ is shown as Algorithm \ref{alg-DPLL-MAPF} (it is based on the DPLL(T) pseudo-code from \cite{DBLP:series/txtcs/KroeningS16}). The algorithm follows the design of {\em conflict-directed clause learning} (CDCL)\footnote{The proper name of the algorithm hence should be CDCL(MAPF), but we follow the notation DPLL(T) used in the literature.} SAT solver \cite{DBLP:conf/iccad/SilvaS96,DBLP:conf/ictai/SilvaS96} into which MAPF consistency check is added when the SAT solver has a partial assignment of Boolean variables at hand (lines 19-20). After a collision is detected the incomplete Boolean model is refined with new collision avoidance clauses (lines 22 and 23) and backtracking based on the analysis of the {\em implication graph} is initiated (lines 13-18). The backtracking phase adds a conflict clause that forbids repeating the conflicting assignment in the future. The high-level that increments the sum-of-costs is the same as in SMT-CBS.

\begin{algorithm}[t]
\begin{footnotesize}
\SetKwBlock{NRICL}{DPLL-MAPF-Fixed ($conflicts,\xi,\Sigma$)}{end} \NRICL{
    $\mathcal{H}(\xi) \gets$ encode-Incomplete$(\mathit{conflicts},\xi,\Sigma)$\\
    \If {propagate-Unit$() = \mathit{UNSAT}$}{
    	\Return {($\mathit{UNSAT}$,$\mathit{conflicts}$)}\\
    }
    \While {$\mathit{TRUE}$}{
        $(x,v) \gets$ assign-Variable()\\
    	  \If {$x = \mathit{NULL}$}{
    	      $\mathit{paths} \gets$ extract-Solution$(\mathit{assignment})$\\
    	      \Return $(\mathit{paths,conflicts})$\\
    	  }
    	  $\mathit{assignment} \gets \mathit{assignment} \cup \{x = v\}$\\
    	  \Repeat{$\mathit{collisions} = \emptyset$}{
    	     \While {propagate-Unit$() = \mathit{UNSAT}$}{
    	         $\mathit{backtrackLevel} \gets$ analyze-Conflict()\\
    	         \If {$\mathit{backtrackLevel} < 0$}{
    	             \Return {($\mathit{UNSAT}$,$\mathit{conflicts}$)}\\
    	         }
    	         \Else{
    	             back-Track($\mathit{backtrackLevel}$)\\
    	         }    	         
    	     }
    	     $\mathit{paths} \gets$ extract-Partial-Solution$(\mathit{assignment})$\\
	     $\mathit{collisions} \gets$ check-Consistency($\mathit{paths}$) /* via $\mathit{DECIDE}_{\mathit{MAPF}}$ */\\
	      \For{each $(a_i,a_j,v,t) \in \mathit{collisions}$}{
                $\mathcal{H}(\xi) \gets \mathcal{H}(\xi) \cup \{\neg \mathcal{X}_v^t(a_i) \vee \neg \mathcal{X}_v^t(a_j)$\}\\
                $\mathit{conflicts} \gets \mathit{conflicts} \cup \{[(a_i,v,t),(a_j,v,t)]\}$
            }
        }
    }
} \caption{Framework of SAT-based MAPF solver} \label{alg-DPLL-MAPF}
\end{footnotesize}
\end{algorithm}

\section{Preliminary Evaluation}
We implemented DPLL(MAPF) in C++ via integrating the MAPF consistency check directly into the Glucose 4 SAT solver \cite{DBLP:journals/ijait/AudemardS18} which effectively changed the SAT solver into a MAPF solver.

Our preliminary tests discovered that the MAPF consistency check is relatively expensive so it cannot be called every time when a partial assignment is available unless we have a more efficient implementation. Hence we check the consistency only at certain stages - our choice is to perform the check when the SAT solver has 1/2 of literals assigned, then 2/3 of literals, and when it is done, denoted DPLL 1/2 3/4. Similarly we used two additional setups: DPLL 1/3 2/3 and DPLL 2/3.

\begin{figure}[t]
  \centering
  \includegraphics[scale=1.2]{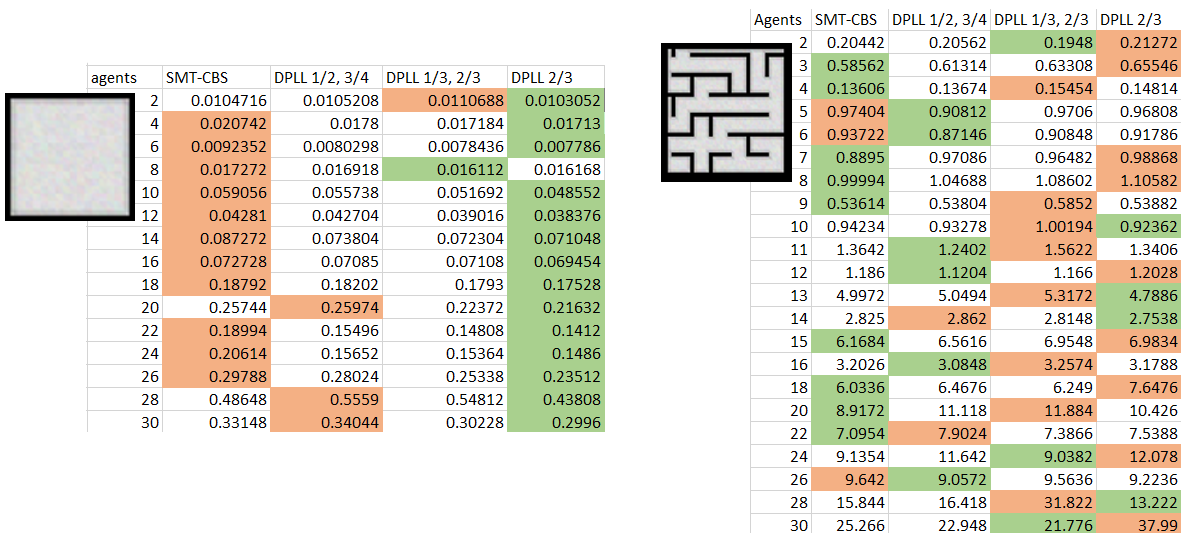}
  \caption{Runtime results for \texttt{empty-16-16} and \texttt{maze-32-32-4} (in seconds).}
  \label{table-empty-maze}
\end{figure}

We compared DPLL(MAPF) with SMT-CBS which has been reimplemented C++ so it shares the code with the DPLL(MAPF) implementation \footnote{Tests were done on a system with Core i7 CPU, 16 GB RAM, under Windows 10.}. The comparison was done on MAPF benchmarks from \texttt{movingai.com} \cite{DBLP:journals/tciaig/Sturtevant12}. Figure \ref{table-empty-maze} show runtime results for two small maps. The results indicate that the DPLL(MAPF) solver has the potential to outperform SMT-CBS especially on small instances where the MAPF consistency check is less expensive.

\section{Conclusion}

We suggested a new innovative step in the SAT-based solving of the multi-agent path finding problem (MAPF) in which the SAT solver and the MAPF reasoning part are closely integrated. The collision resolution in solutions obtained from partial assignments of incomplete Boolean models is integrated directly in the SAT solver's search loop which effectively turns the SAT solver into a MAPF solver. We demonstrated in a preliminary evaluation that DPLL(MAPF) has a potential to outperform previous SAT-based MAPF solvers.

The impact of DPLL(MAPF) is reaching further as we can use the same framework for implementing DPLL(MAPF$_R$), that is a SAT-based solver for the continuous variant of MAPF (MAPF$_R$) \cite{DBLP:conf/ijcai/AndreychukYAS19}.

Even DPLL(classical planning) seems to be possible and could represent a next step for SAT-based classical planners \cite{DBLP:books/daglib/0014222,DBLP:conf/socs/FroleyksBS19} especially in domains involving multiple interacting agents \cite{DBLP:conf/iat/DimopoulosM06,DBLP:journals/kbs/DimopoulosHM12}.


\section{Acknowledgement}

This research has been supported by GA\v{C}R - the Czech Science Foundation, under the grant number 19-17966S, and by Student Summer Research Program 2020 of FIT CTU in Prague (V\'{y}Let 2020).

\bibliographystyle{splncs04}
\bibliography{references}

\end{document}